\pdfoutput=1

\documentclass[11pt]{article}

\usepackage[final]{acl}

\usepackage{times}
\usepackage{latexsym}

\usepackage[T1]{fontenc}

\usepackage[utf8]{inputenc}

\usepackage{microtype}

\usepackage{inconsolata}

\usepackage{graphicx}

\usepackage{amsmath}
\usepackage{amssymb}
\usepackage{mathtools}
\usepackage{amsthm}
\usepackage{pifont}
\usepackage{subcaption}

\usepackage{multirow}
\usepackage{booktabs}
\usepackage{soul,xcolor,todonotes}
\definecolor{kmycolor}{rgb}{0.8,0.2,0.4}
\definecolor{forestgreen}{rgb}{0.0, 0.5, 0.0}
\newtheorem{proposition}{Proposition}

\title{V-DPO: Mitigating Hallucination in Large Vision Language Models via Vision-Guided Direct Preference Optimization}

\author{Yuxi Xie \quad Guanzhen Li \quad Xiao Xu \quad Min-Yen Kan \\
  National University of Singapore \\
  \texttt{\{xieyuxi, guanzhen, xuxiao\}@u.nus.edu} \quad \texttt{knmnyn@nus.edu.sg}
}

\begin{document}
\maketitle
\begin{abstract}
Large vision--language models (LVLMs) suffer from hallucination, 
resulting in misalignment between the output textual response and the input visual content. Recent research indicates that the over-reliance on the Large Language Model (LLM) backbone, as one cause of the LVLM hallucination, inherently introduces bias from language priors, leading to insufficient context attention to the visual inputs. 

We tackle this issue of hallucination by mitigating such over-reliance through preference learning. We propose Vision-guided Direct Preference Optimization (V-DPO) 
to enhance visual context learning at training time. To interpret the effectiveness and generalizability of V-DPO on different types of training data, we construct a synthetic dataset containing both response- and image-contrast preference pairs, compared against existing human-annotated hallucination samples. Our approach achieves significant improvements
compared with baseline methods 
across various hallucination benchmarks. Our analysis indicates that V-DPO excels in learning from image-contrast preference data, demonstrating its superior ability to elicit and understand nuances of visual context.
Our code is publicly available at \href{https://github.com/YuxiXie/V-DPO}{https://github.com/YuxiXie/V-DPO}.
\end{abstract}

\section{Introduction}\label{sec:intro}
\begin{figure}[t]
  \centering
  \begin{subfigure}[b]{0.42\columnwidth}
      \includegraphics[width=\textwidth]{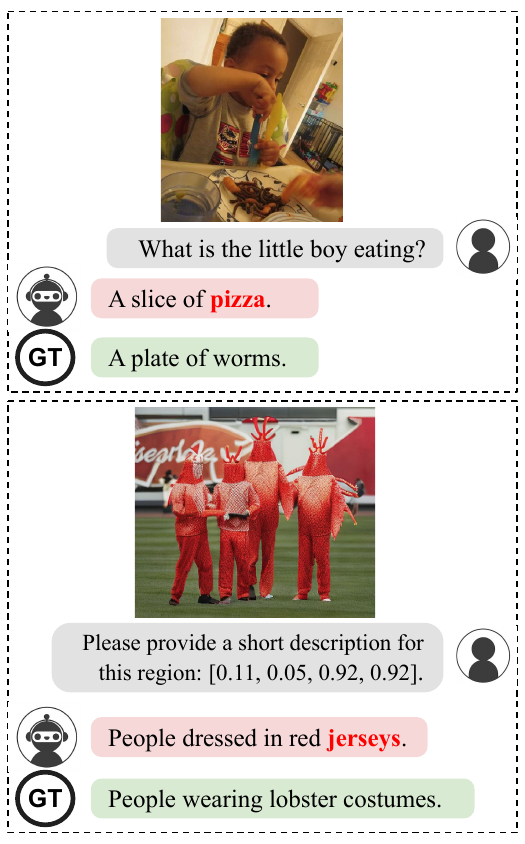}
      \caption{Hallucination examples on unconventional visual contents.}
      \label{fig:exp}
  \end{subfigure}
  \hfill
  \begin{subfigure}[b]{0.56\columnwidth}
      \includegraphics[width=\textwidth]{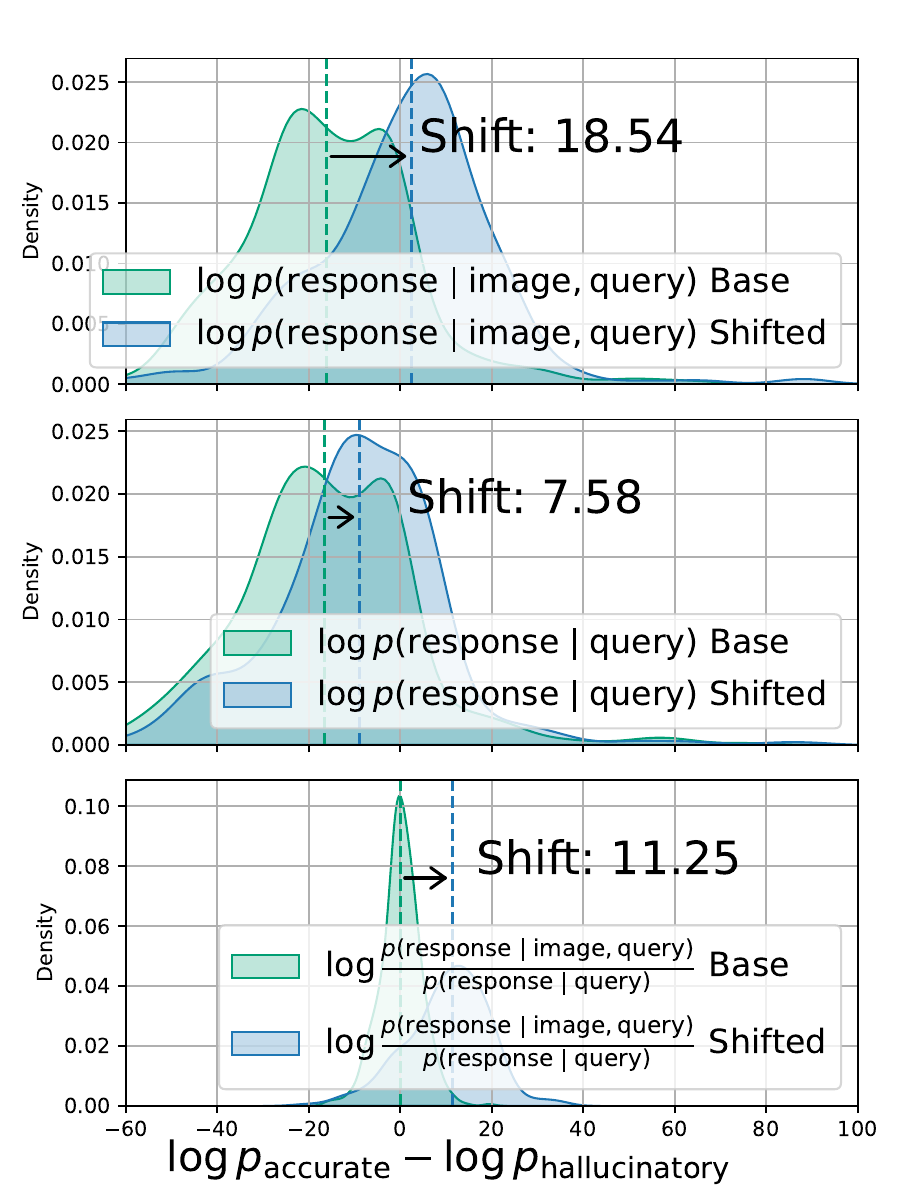}
      \caption{Distribution gaps.}
      \label{fig:gap}
  \end{subfigure}
  \caption{(a) Hallucination examples in visual question answering and region descriptions and (b) the model discriminative ability on the accurate and hallucinatory samples represented by difference in log-likelihoods.}
\end{figure}

%
%
%
Recent advancements in Large Language Models (LLMs)~\citep{DBLP:conf/nips/BrownMRSKDNSSAA20,DBLP:journals/jmlr/ChowdheryNDBMRBCSGSSTMRBTSPRDHPBAI23,DBLP:journals/corr/abs-2302-13971,vicuna2023,DBLP:journals/corr/abs-2303-08774} have catalyzed the evolution of Large Vision--Language Models (LVLMs)~\citep{DBLP:conf/nips/LiuLWL23a,DBLP:journals/corr/abs-2310-03744,DBLP:conf/nips/Dai0LTZW0FH23,DBLP:journals/corr/abs-2312-11805} in understanding and reasoning across visual and textual modalities. Despite their impressive performance on various vision--language tasks, existing LVLMs still struggle with the issue of \textit{hallucination}, where the model outputs are not factually grounded in the input visual contents~\citep{DBLP:conf/emnlp/RohrbachHBDS18, DBLP:conf/emnlp/LiDZWZW23, DBLP:conf/aaai/GunjalYB24,DBLP:journals/corr/abs-2402-00253}. Hallucination in LVLMs refers to non-existing or erroneous descriptions of visual contents, such as objects, attributes, and relationships, which is especially challenging to understanding unconventional images, as shown in Figure~\ref{fig:exp}.

The phenomenon of hallucination in LVLMs can be attributed to the integration of pre-trained LLMs in the architecture. 
Recent works reveal that this issue is closely tied to insufficient context attention, where the model prioritizes language patterns and focuses on partial tokens rather than fully grounding the generated content in both visual and textual context~\citep{DBLP:journals/corr/abs-2311-07362,DBLP:conf/aaai/WangWHPZZDLLWH24}. 
To mitigate the over-reliance on language priors, many efforts have been devoted to decoding optimization with penalties on over-trust candidates~\citep{DBLP:journals/corr/abs-2311-17911} or a focus on visual uncertainty~\citep{DBLP:journals/corr/abs-2403-00425}. However, these methods require increased inference time and specific infrastructure designs~\citep{DBLP:journals/corr/abs-2311-07362}, obstructing their generalizability and scalability across diverse data domains and sizes. In contrast, our study explores training strategies to alleviate the over-reliance on language priors via preference learning, enhancing visual understanding to mitigate hallucination in LVLMs.

Given the difference in the likelihoods between accurate and hallucinatory samples on vision-conditioned $p(\mathrm{response}\mid\mathrm{image},\mathrm{query})$ and textual-only $p(\mathrm{response}\mid\mathrm{query})$ distributions, Figure~\ref{fig:gap} illustrates the shifts of this difference after aligning the model with hallucination-free data via preference learning. Before alignment, the textual-only distributions dominate the model decision on determining accurate samples as preferred compared to hallucinatory ones, reflected by the distributions (in green) of the same shape for both probabilities. This dominance in pairwise preference illustrates the over-reliance on language priors in LVLMs, which is especially crucial for unseen images in training (\textit{e.g.}, Figure~\ref{fig:exp}), limiting the model generalizability across different data. 
Motivated by this challenge, we propose Vision-guided Direct Preference Optimization (V-DPO), a vision-specific variant of Direct Preference Optimization (DPO)~\citep{DBLP:conf/nips/RafailovSMMEF23}, to employ visual guidance during preference learning for hallucination mitigation in LVLMs. 
We adapt Classifier-Free Guidance (CFG)~\citep{DBLP:journals/corr/abs-2207-12598} to integrate the visual guidance into the optimization target, inspired by its effectiveness in improving the specificity of model generations tailored for specific contents~\citep{DBLP:journals/corr/abs-2306-17806, DBLP:conf/iccv/KornblithLWN23}. 
To assess the generalizability of V-DPO, especially on unconventional contents, we construct a synthetic dataset containing both response-contrast and image-contrast preference pairs, compared against existing human-annotated preferences such as RLHF-V~\citep{DBLP:journals/corr/abs-2312-00849}. Our approach exhibits significant and stable performance improvements through extensive experiments on various hallucination benchmarks. Further analysis of the distribution shifts from training demonstrates the effectiveness of V-DPO in mitigating the over-reliance on language priors on both image- and response-contrast data.

\section{Related Work}~\label{sec:related-work}
Hallucination has emerged as a significant challenge to model reliability and generalizability in LVLM development. To alleviate hallucinated content, existing works can be divided as following two directions. The first focuses on post-processing approaches, including post-hoc corrections~\citep{DBLP:journals/corr/abs-2310-00754,DBLP:journals/corr/abs-2310-16045,DBLP:journals/corr/abs-2311-07362} and specialized decoding~\citep{DBLP:journals/corr/abs-2311-17911,DBLP:journals/corr/abs-2403-00425}. 
However, these methods often require increased inference time, obstructing their generalizability and scalability across diverse data domains and sizes~\citep{DBLP:journals/corr/abs-2404-18930}. 

The second line of work attempts to collect hallucination-aware data to mitigate hallucination in LVLMs through preference optimization leaning toward hallucination-free outputs. For example, \citet{DBLP:journals/corr/abs-2309-14525} and \citet{DBLP:journals/corr/abs-2312-00849} adapt the Reinforcement Learning from Human Feedback (RLHF) and Direct Preference Optimization (DPO) paradigms in LLMs, respectively, to align LVLMs with hallucination-aware human preferences. \citet{DBLP:journals/corr/abs-2311-16839} and \citet{sarkar2024mitigating} propose data augmentation pipelines to construct (accurate, hallucinatory) preference pairs for contrastive tuning. Our work mitigates hallucination in the context of preference optimization with not only augmented data including both response- and image-contrast preference pairs, but also a vision-specific optimization target to enhance visual understanding.
\section{Background and Motivations}~\label{sec:preliminary}
We explore strategies to enhance visual understanding in LVLM preference optimization. Our framework starts from a supervised fine-tuned (SFT) model, obtained by jointly training a visual encoder and a pre-trained LLM via visual instruction tuning~\citep{DBLP:conf/nips/LiuLWL23a}. Specifically, we incorporate visual guidance by integrating Classifier-Free Guidance (CFG) into vanilla DPO.

\subsection{Preference Optimization for LVLMs}~\label{sec:dpo}
We consider a policy LVLM $\pi_{\theta}$ parameterized by $\theta$. For a vision-conditioned text generation task, given an input image $v\sim\mathcal{I}$ and a textual query $x\sim\mathcal{P}$, we optimize for the KL-constrained reward maximization objective:
\begin{equation}
    \label{eq:dpo-obj}
    \begin{aligned}
         &\max_{\pi}\mathbb{E}_{(v,x)\sim\mathcal{I}\times\mathcal{P},y\sim\pi}\Bigl[r(v,x,y) \\
         - &\beta\mathbb{D}_{\mathrm{KL}}\left[\pi(y\mid v,x)\parallel\pi_{\mathrm{ref}}(y\mid v,x)\right]\Bigr]
    \end{aligned}
\end{equation}
under reward function $r(v,x,y)$ and reference model $\pi_{\mathrm{ref}}$. DPO solves the optimal policy as:
\begin{equation}
    \label{eq:dpo-policy}
    \begin{aligned}
        \pi_r(y\mid v,x) = & \frac{\pi_{\mathrm{sft}}(y\mid v,x)\exp\Bigl(\frac{1}{\beta}r(v,x,y)\Bigl)}{Z(v,x)}
    \end{aligned}
\end{equation}
for all image--query pairs $(v,x)\sim\mathcal{I}\times\mathcal{P}$, where $Z(v,x) = \sum_y{\pi_{\mathrm{sft}}(y\mid v,x)\exp{\Bigl(\frac{1}{\beta}r(v,x,y)\Bigl)}}$ is the partition function. 

Rearranging Eq.~\ref{eq:dpo-policy}, we get the ground-truth reward model with the corresponding optimal policy. Given a response-contrast preference dataset $\mathcal{D}_y = \{v^{(k)}, x^{(k)}, y_w^{(k)}, y_l^{(k)}\}_{k=1}^N$ where $y_w$ is preferred over $y_l$, DPO uses Bradley–Terry model~\citep{bradley1952rank} to derive the objective as:
\begin{equation}
    \label{eq:dpo-loss}
    \begin{aligned}
        \mathcal{L}_{\mathrm{DPO}}^y(\pi_{\theta}; \pi_{\mathrm{ref}}) = -\mathbb{E}_{(v,x,y_w,y_l)\sim\mathcal{D}_y}\log\sigma(\beta u_{\pi_{\theta}}^{y_w, y_l})
    \end{aligned}
\end{equation}
where 
$u_{\pi_{\theta}}^{y_w, y_l} = \log\frac{\pi_{\theta}(y_w\mid v,x)}{\pi_{\mathrm{ref}(y_w\mid v,x)}} - \log\frac{\pi_{\theta}(y_l\mid v,x)}{\pi_{\mathrm{ref}(y_l\mid v,x)}}$ indicates the implicit reward corresponding to  $\pi_{\theta}$.


Enlightened by \textit{contrast sets}~\citep{DBLP:conf/emnlp/0001ABBBCDDEGGH20, DBLP:journals/corr/abs-2309-16155}, we construct an image-constrast dataset $\mathcal{D}_{v} = \{v_w^{(k)}, v_l^{(k)}, x^{(k)}, y^{(k)}\}_{k=1}^M$ to enhance visual understanding. With $u_{\pi_{\theta}}^{v_w, v_l} = \log\frac{\pi_{\theta}(y\mid v_w,x)}{\pi_{\mathrm{ref}}(y\mid v_w,x)} - \log\frac{\pi_{\theta}(y\mid v_l,x)}{\pi_{\mathrm{ref}}(y\mid v_l,x)}$, we have:
\begin{equation}
    \label{eq:dpo-loss2}
    \begin{aligned}
        \mathcal{L}_{\mathrm{DPO}}^{v}(\pi_{\theta}; \pi_{\mathrm{ref}}) = -\mathbb{E}_{(v_w, v_l,x,y)\sim\mathcal{D}_v}\log\sigma(\beta u_{\pi_{\theta}}^{v_w, v_l})
    \end{aligned}
\end{equation}

\subsection{Classifier-Free Guidance in LLMs}~\label{sec:cfg}
CFG was originally proposed in the context of conditioned diffusion models~\citep{DBLP:conf/nips/DhariwalN21}. Given a noisy image $y$ and a class condition $c$, the model predicts probability likelihood $\hat{p}$ for the conditioned step-wise sample $\hat{\pi}_{\theta}(y\mid c)\varpropto \pi_{\theta}(y)\cdot \pi_{\phi}(c\mid y)^{\gamma}$, where $\gamma > 0$ controls the guidance strength from the classifier $\pi_{\phi}$. \citet{DBLP:journals/corr/abs-2207-12598} observe that the guidance can be offered without a classifier:
\begin{equation}
    \label{eq:cfg}
    \begin{aligned}
        \hat{\pi}_{\theta}(y\mid c)\varpropto \pi_{\theta}(y)\cdot \pi_{\theta}(c\mid y)^{\gamma} \varpropto \frac{\pi_{\theta}(y\mid c)^{\gamma}}{\pi_{\theta}(y)^{\gamma - 1}}
    \end{aligned}
\end{equation}

Given a textual completion $\mathbf{y}=\{y_i\}_{i=1}^N$ and a conditional prompt or image $c$, we can extend CFG to autoregressive models as $\hat{\pi}_{\theta}(\mathbf{y}\mid c) \varpropto \frac{\pi_{\theta}(\mathbf{y}\mid c)^{\gamma}}{\pi_{\theta}(\mathbf{y})^{\gamma-1}} \varpropto \prod_{i=1}^N\frac{\pi_{\theta}(y_i\mid y_{< i}, c)^{\gamma}}{\pi_{\theta}(y_i\mid y_{< i})^{\gamma - 1}}$. Previous works show that CFG increases the specificity of the generation to be more pertinent toward the prompt~\citep{DBLP:journals/corr/abs-2306-17806} or image~\citep{DBLP:conf/iccv/KornblithLWN23}.
Enlightened by this insight, we apply CFG in LVLM preference optimization to enhance the importance of visual context. This employment is non-trivial considering the dynamics in the training process, which we will detail next.

\section{Vision-Guided Preference Learning}~\label{sec:method}
%
%
%
%
%
In this work, we focus on mitigating hallucinations in LVLMs caused by insufficient context attention to visual information. We propose Vision-guided Direct Preference Optimization (V-DPO) to enhance visual understanding on both response- and image-contrast preference data. 

\begin{figure*}[t]
    \centering
    \includegraphics[width=\textwidth]{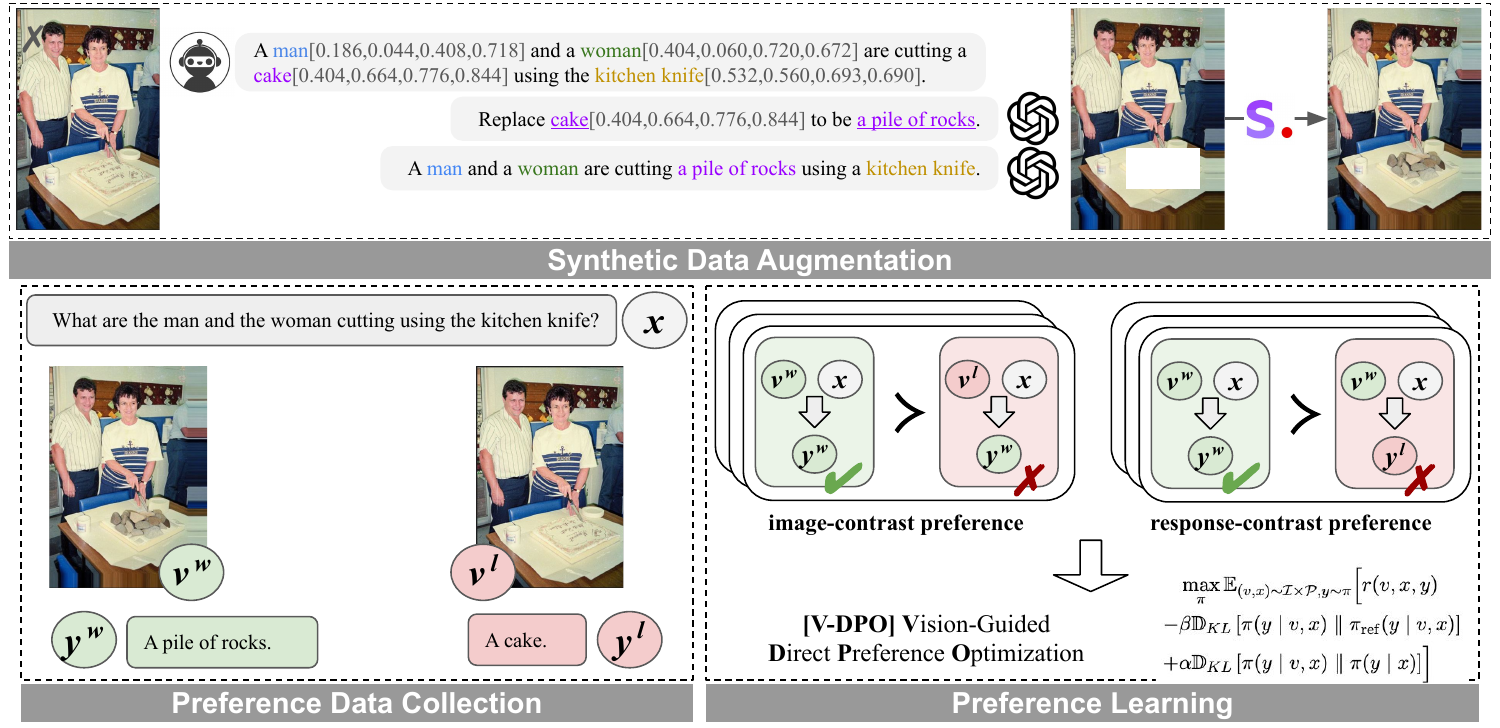}
    \caption{Outline of our preference data construction and vision-guided preference learning framework. In the stage of Synthetic Data Augmentation, we utilize LVLMs, LLMs, and Stable Diffusion to manipulate images automatically. We formulate the generated samples into image- and response-contrast pairs for preference learning via our Vision-guided DPO approach.}
    \label{fig:pipeline}
\end{figure*}

\subsection{Vision-Guided DPO}~\label{sec:vdpo}
Our V-DPO approach builds on the insight that CFG-modified distribution produces more condition-specific generation than vanilla decoding. As we will detail next, our core contribution originates from a vision-specific term in the reward maximization objective of DPO.

\paragraph{V-DPO Objective.}
We start with the definition of visual guidance in the context of LVLMs. Following Eq.~\ref{eq:cfg}, we apply CFG to vision-conditioned text generation:
\begin{equation}
    \label{eq:cfg-vlm}
    \begin{aligned}
        \hat{\pi}_{\theta}(y\mid v,x) \varpropto \pi_{\theta}(y\mid x)\left(\frac{\pi_{\theta}(y\mid v, x)}{\pi_{\theta}(y\mid x)}\right)^{\gamma}
    \end{aligned}
\end{equation}
where $\frac{\pi_{\theta}(y\mid v, x)}{\pi_{\theta}(y\mid x)}$ is the guidance from the visual context $v$ to increase the specificity of the response $y$ toward the image, given the input query $x$. We integrate this term as an additional target to optimize in Eq.~\ref{eq:dpo-obj}. Our result vision-enhanced reward maximization objective is then:
\begin{equation}
    \label{eq:vdpo-obj}
    \begin{aligned}
        &\max_{\pi}\mathbb{E}_{(v,x)\sim\mathcal{I}\times\mathcal{P},y\sim\pi}\Bigl[r(v,x,y) \\
        - &\beta\mathbb{D}_{\mathrm{KL}}\left[\pi(y\mid v,x)\parallel\pi_{\mathrm{ref}}(y\mid v,x)\right] \\
        + & \alpha\mathbb{D}_{\mathrm{KL}}\left[\pi(y\mid v,x)\parallel\pi(y\mid x)\right]\Bigr]
    \end{aligned}
\end{equation}
where $\alpha > 0$ controls the weight of the visual guidance to optimize. Solving the optimal solution $\pi_r$ to the above objective, we have:
\begin{equation}
    \label{eq:vdpo-policy}
    \begin{aligned}
        & \pi_r(y\mid v,x)^{\gamma} / \pi_r(y\mid x)^{\gamma - 1} \\
        = & \pi_r(y\mid v,x) \left(\frac{\pi_r(y\mid v,x)}{\pi_r(y\mid x)}\right)^{\gamma - 1} \\
        \varpropto & \frac{1}{Z(v,x)}\pi_{\mathrm{sft}}(y\mid v,x)\exp\Bigl(\frac{1}{\beta}r(v,x,y)\Bigl) 
    \end{aligned}
\end{equation}
where $\gamma = 1 - \frac{\alpha}{\beta}$. Unlike inference-time CFG, we decrease $\gamma < 1$; \textit{i.e.}, increasing $\alpha > 0$, to strengthen the guidance of visual context during training. We detail the complete derivations in Appendix~\ref{app:theory}. Although only a proportional relationship holds here (as $\pi_r(y\mid v,x)^{\gamma} / \pi_r(y\mid x)^{\gamma - 1}$ is an unnormalized probability distribution), we can still obtain the reward difference of a preference pair using the Bradley--Terry model. Similar to Eqs.~\ref{eq:dpo-loss} and \ref{eq:dpo-loss2}, we derive our policy objective as:
\begin{equation}
    \label{eq:vdpo-loss}
    \begin{aligned}
        \mathcal{L}_{\mathrm{VDPO}}(\pi_{\theta}; \pi_{\mathrm{ref}}) = -\mathbb{E}_{(w, l)\sim\mathcal{D}}\log\sigma(\beta u_{\pi_{\theta}}^{w, l})
    \end{aligned}
\end{equation}
where $\mathcal{D}=\mathcal{D}_y\cup\mathcal{D}_v$ and $u_{\pi_{\theta}}^{w,l} = f_{\theta}^w - f_{\theta}^l$, using the shorthand $f_{\theta}(v, x, y) = \log\frac{\pi_{\theta}(y\mid v,x)\hat{\varphi}_{\theta}(v,x,y)}{\pi_{\mathrm{ref}(y\mid v.x)}}$ with $\varphi_{\theta}(v,x,y) = \left(\frac{\pi_{\theta}(y\mid v,x)}{\pi_{\theta}(y\mid x)}\right)^{\gamma - 1}$ controlling the strength of visual guidance. 

\paragraph{Implementation of Visual Guidance.}
In Eq.~\ref{eq:vdpo-loss}, we disable gradient backpropagation on ${\varphi}_{\theta}(v,x,y)$ to maintain a stable textual-only distribution $\pi_{\theta}(\cdot\mid x)$ during training. This aims to provide a reliable reference value to calculate the visual guidance. We further discuss the choice of $\hat{\pi}(\cdot\mid x)$ in Section~\ref{sec:ablation}. Following the implementation of \citet{DBLP:conf/nips/LiuLWL23a}, we pass zeroes in place of the conditioning visual context to get the textual-only distribution:
\begin{equation}
    \label{eq:noimage}
    \hat{\pi}_{\theta}(\cdot\mid x) = \hat{\pi}_{\theta}(\cdot\mid\mathbf{0}, x)
\end{equation}

With the integration of visual guidance, we modify $\pi_{\theta}(y\mid v,x)$ in vanilla DPO to be a non-normalized probability distribution, $\pi_{\theta}(y\mid v,x)\hat{\varphi}_{\theta}(v,x,y)$. Empirically, this can progressively decrease the effect of visual guidance as the visual-conditioned and unconditioned distributions diverge from each other through training. To mitigate this problem, we follow \citet{DBLP:conf/iccv/KornblithLWN23} to normalize it as:
\begin{equation}
    \label{eq:norm-cfg}
    \begin{aligned}
        & \pi_{\theta}(\cdot\mid v,x)\hat{\varphi}_{\theta}(v,x,\cdot) \\
        \varpropto & \phi\left(h_{\theta}(v, x) + (\gamma - 1)\left(\hat{h}_{\theta}(v, x) - \hat{h}_{\theta}(\mathbf{0}, x)\right)\right)
    \end{aligned}
\end{equation}
where $h_{\theta}$ are the generated logits and $\phi(\cdot)$ is the softmax function for normalization. Note that since the increase of divergence between the distribution $\pi_{\theta}(\cdot\mid v,x)$ and $\pi_{\theta}(\cdot\mid x)$ can lead to a larger exponential sum in softmax, the normalization thus gradually inflates the effect of visual guidance during training. We analyze the potential impacts of the guidance inflation in Section~\ref{sec:analysis}.

\subsection{Constructing Contrast Images}~\label{sec:ic}
As discussed in Section~\ref{sec:dpo}, we augment the preference data with image-contrast pairs to enhance visual understanding via preference learning. The construction of contrastive image pairs aims to bolster the visual understanding ability to discern nuanced visual differences between similar images. Specifically, we manipulate images by replacing conventional items with unconventional ones, considering the limited capability of LVLMs to understand weird images~\citep{DBLP:conf/iccv/GuettaBHSES023}.
This section details the automatic construction process we use to collect image-contrast preference data. 

\paragraph{Proposing Replacement Elements.}
Given an image from an existing dataset, we extract object-level information using LVLMs and generate detailed captions with objects grounded in respective positions in the image. Based on the layout-grounded descriptions, we employ LLMs to propose replacements for visual elements, thereby creating unexpected scenarios by leveraging their imaginative capability~\citep{DBLP:conf/emnlp/Gomez-Rodriguez23a}. Figure~\ref{fig:pipeline} shows an example element replacement proposed by ChatGPT. To enhance the interpretability of this automatic process, we require LLMs to supply a reasonable explanation of the replacement's unexpectedness (\textit{cf.} Appendix~\ref{app:prompts} for prompts and examples). We collect multiple replacements for each image, which are used to guide image generation next.

\paragraph{Image Editing and Filtering.}
Given a designated region in a source image, we use a generative model to edit via image inpainting~\citep{DBLP:conf/cvpr/LugmayrDRYTG22}. Particularly, we utilize Denoising Diffusion Probabilistic Models (DDPMs) as the image inpainter, considering their superior generation quality~\citep{DBLP:conf/nips/DhariwalN21}. Empirically, the imperfections of the LLM and the generative model can result in a significant distribution gap between the generated images and the original real ones, introducing noise and bias into the synthetic data. To address this issue, we use CLIPScore~\citep{DBLP:conf/emnlp/HesselHFBC21} to refine our data by filtering out edits that do not align well with the corresponding replacement prompts. Specifically, we approve an edited image $v_i$ only if it achieves the highest CLIPScore with the intended textual prompt $c_i$ in comparison with similar text--image pairs generated in our pipeline:
\begin{equation}
    \label{eq:clipscore}
    \begin{aligned}
        c_i = & \arg\max_{c}\mathrm{CLIPScore}(c, v_i) \\
        v_i = & \arg\max_{v}\mathrm{CLIPScore}(c_i, v)
    \end{aligned}
\end{equation}

Finally, we combine our image-contrast pairs with conventional response-contrast ones to construct our preference data for V-DPO. See Appendix~\ref{app:data} for a full construction pipeline for different types of preference data.

\section{Experiments}~\label{sec:exp}
We now assess V-DPO across various multimodal hallucination benchmarks. To interpret how V-DPO improves visual understanding, we compare performance using various preference data. Specifically, unlike previous studies focusing on performance improvement using specific data, this work aims to demonstrate the effectiveness and generalizability of V-DPO across different training datasets and benchmarks for fair comparison.

\subsection{Setup}
We choose \textsc{LLaVA-v1.5-7B}~\citep{DBLP:journals/corr/abs-2310-03744} as our initial SFT model and conduct preference learning with full fine-tuning. Our synthetic augmented data contains 5K response- and image-contrast preference pairs, compared against the human-annotated response-contrast data RLHF-V (5K)~\citep{DBLP:journals/corr/abs-2312-00849} of equal size. In Appendix~\ref{app:exp}, we further conduct extended experiments on \textsc{LLaVA-v1.6-7B} to demonstrate the generalizability of V-DPO.\looseness=-1

\paragraph{Benchmarks.}
We evaluate our approach on four hallucination benchmarks: (1) POPE~\citep{DBLP:conf/emnlp/LiDZWZW23} on object hallucination with discriminative tasks; (2) AMBER~\citep{DBLP:journals/corr/abs-2311-07397} containing both generative and discriminative tasks on object, attribute, and relation hallucination; (3) HallusionBench~\citep{DBLP:journals/corr/abs-2310-14566} assessing visual illusion and knowledge hallucination with systematically structured discriminative tasks; and (4) MMHal-Bench~\citep{DBLP:journals/corr/abs-2309-14525} covering different question types and object topics. We also conduct general-purpose evaluation on MMBench~\citep{DBLP:conf/iiswc/XuHLLHZNSTXCG23} across various multimodal tasks in Appendix~\ref{app:mmbench}.

\paragraph{Baselines.}
We compare our method against the initial SFT model and vanilla DPO as the fundamental and strengthened baselines, respectively. 
We also consider Hallucination-Aware Direct Preference Optimization (HA-DPO)~\citep{DBLP:journals/corr/abs-2311-16839} as a variant of DPO baseline trained on 16K style-consistent hallucination sample pairs.

\subsection{Main Results}~\label{sec:results}
We compare V-DPO with vanilla DPO methods across various hallucination benchmarks to show the effectiveness and stability of our approach.

\begin{table*}
    \centering
    \small
    \begin{minipage}[b]{0.58\textwidth}
        \centering
        \begin{tabular}{lccccc}
            \hline
            \multirow{2}{*}{\textbf{Approach}} & \multicolumn{4}{c}{\textbf{F1 Score}}& \multirow{3}{*}{\shortstack{\textbf{Yes}\\\textbf{Ratio}}} \\
            \cmidrule(lr){2-5}
            & \textbf{F1}\textsubscript{R $\uparrow$} & \textbf{F1}\textsubscript{P $\uparrow$} & \textbf{F1}\textsubscript{A $\uparrow$} & \textbf{F1}\textsubscript{$\uparrow$} & \\
            \hline
            SFT & $89.69$ & $86.83$ & $81.80$ & $85.98$ & $54.20$ \\
            HA-DPO & $\mathbf{90.25}$ & $87.81$ & $82.54$ & $86.87$ & $51.03$ \\
            \hline
            \multicolumn{6}{c}{\textbf{Synthetic Augmented Data}} \\
            DPO & $88.34$ & $87.05$ & $83.96$ & $86.42$ & $44.22$ \\
            V-DPO & $89.57$ & $87.62$ & $83.77$ & $86.92$\textcolor{forestgreen}{$_{\uparrow 0.94}$} & $47.43$ \\
            \hline
            \multicolumn{6}{c}{\textbf{RLHF-V}} \\
            DPO & $89.69$ & $87.81$ & $84.03$ & $87.12$ & $47.88$ \\
            V-DPO & $89.90$ & $\mathbf{87.91}$ & $\mathbf{84.05}$ & $\mathbf{87.22}$\textcolor{forestgreen}{$_{\uparrow 1.24}$} & $48.66$ \\
            \hline
        \end{tabular}
        \caption{Result comparison (F1 score) on POPE including splits of random (R), popular (P), and adversarial (A) scenarios. We report Yes Ratio ($\%$) to compare the biased tendency of different models.}
        \label{tab:pope}
    \end{minipage}
    \hfill
    \begin{minipage}[b]{0.4\textwidth}
        \centering
        \small
        \begin{tabular}{lccc}
            \hline
            \multirow{2}{*}{\textbf{Approach}} & \multicolumn{3}{c}{\textbf{Accuracy}} \\
            \cmidrule(lr){2-4}
            & \textbf{qAcc}\textsubscript{$\uparrow$} & \textbf{fAcc}\textsubscript{$\uparrow$} & \textbf{aAcc}\textsubscript{$\uparrow$} \\
            \hline
            SFT & $13.19$ & $20.23$ & $48.16$ \\
            \hline
            \multicolumn{4}{c}{\textbf{Synthetic Augmented Data}} \\
            DPO & $21.97$ & $20.52$ & $\mathbf{55.52}$ \\
            V-DPO & $\mathbf{22.20}$\textcolor{forestgreen}{$_{\uparrow 9.01}$} & $\mathbf{21.10}$ & $55.31$ \\
            \hline
            \multicolumn{4}{c}{\textbf{RLHF-V}} \\
            DPO & $16.70$ & $20.81$ & $51.31$ \\
            V-DPO & $17.36$\textcolor{forestgreen}{$_{\uparrow 4.17}$} & $19.94$ & $51.63$ \\
            \hline
        \end{tabular}
        \caption{Results on HallusionBench. qAcc and fAcc assess the accuracy of answering a question and understanding a figure, paired with different images and questions, respectively.}
        \label{tab:hb}
    \end{minipage}
\end{table*}
\begin{table*}
  \centering
  \small
  \begin{tabular}{l cccc cccc c}
    \hline
    \multirow{2}{*}{\textbf{Approach}} & \multicolumn{4}{c}{\textbf{Generative}} & \multicolumn{4}{c}{\textbf{Discriminative}} & \multirow{3}{*}{\shortstack{\textbf{AMBER}\\\textbf{Score}$_{\uparrow}$}} \\
    \cmidrule(lr){2-5} \cmidrule(lr){6-9}
    & \textbf{CHAIR$_{\downarrow}$} & \textbf{Cover$_{\uparrow}$} & \textbf{Hal$_{\downarrow}$} & \textbf{Cog$_{\downarrow}$} & \textbf{F1\textsubscript{E}$_{\uparrow}$} & \textbf{F1\textsubscript{A}$_{\uparrow}$} & \textbf{F1\textsubscript{R}$_{\uparrow}$} & \textbf{F1$_{\uparrow}$} &  \\
    \hline
    SFT & $7.8$ & $51.0$ & $36.4$ & $4.2$ & $64.6$ & $65.6$ & $62.4$ & $74.7$ & $83.5$ \\
    HA-DPO & $6.7$ & $49.8$ & $30.9$ & $3.3$ & $88.1$ & $66.1$ & $\mathbf{68.8}$ & $78.1$ & $85.7$ \\
    \hline
    \multicolumn{10}{c}{\textbf{Synthetic Augmented Data}} \\
    DPO & $7.3$ & $50.2$ & $33.6$ & $3.7$ & $\mathbf{95.2}$ & $75.1$ & $60.9$ & $83.1$ & $87.9$ \\
    V-DPO (Ours) & $6.6$\textcolor{forestgreen}{$_{\downarrow 1.2}$} & $49.1$\textcolor{red}{$_{\downarrow 1.9}$} & $30.8$\textcolor{forestgreen}{$_{\downarrow 5.6}$} & $3.1$\textcolor{forestgreen}{$_{\downarrow 1.1}$} & $95.1$ & $\mathbf{76.1}$ & $61.1$ & $\mathbf{83.5}$\textcolor{forestgreen}{$_{\uparrow 8.8}$} & $\mathbf{88.4}$\textcolor{forestgreen}{$_{\uparrow 4.9}$} \\
    \hline
    \multicolumn{10}{c}{\textbf{RLHF-V}} \\
    DPO & $5.7$ & $49.7$ & $\mathbf{27.3}$ & $\mathbf{2.6}$ & $90.7$ & $72.6$ & $64.6$ & $80.9$ & $87.6$ \\
    V-DPO (Ours) & $\mathbf{5.6}$\textcolor{forestgreen}{$_{\downarrow 2.2}$} & $49.7$\textcolor{red}{$_{\downarrow 1.3}$} & $\mathbf{27.3}$\textcolor{forestgreen}{$_{\downarrow 9.1}$} & $2.7$\textcolor{forestgreen}{$_{\downarrow 1.5}$} & $91.5$ & $73.7$ & $64.1$ & $81.6$\textcolor{forestgreen}{$_{\uparrow 5.9}$} & $88.0$\textcolor{forestgreen}{$_{\uparrow 4.5}$} \\
    \hline
  \end{tabular}
  \caption{Result comparison on AMBER. For generative tasks, we use CHAIR~\citep{DBLP:conf/emnlp/RohrbachHBDS18}, Cover (coverage of ground-truth objects), Hal (hallucination rate), and Cog (Cognition) as evaluation metrics. We report the performance of discriminative tasks using F1 scores, including splits of existence (E), attribute (A), and relation (R). The holistic AMBER Score~\citep{DBLP:journals/corr/abs-2311-07397} is calculated by $(100 - \mathrm{CHAIR} + \mathrm{F1}) / 2$. We compare with HA-DPO~\citep{DBLP:journals/corr/abs-2311-16839} backboned with the same SFT model, LLaVA-v1.5-7B~\citep{DBLP:journals/corr/abs-2310-03744}.}
  \label{tab:amber}
\end{table*}

\paragraph{POPE.} Table~\ref{tab:pope} compares model performance (F1 score) and tendency to answer ``yes'' (Yes Ratio) on POPE. V-DPO outperforms the SFT and vanilla DPO baselines on random sets and more challenging tasks such as the adversarial scenario. Furthermore, V-DPO significantly increases the F1 scores from $85.98$ to $86.92$ and $87.22$ trained on synthetic and human-annotated data, respectively, with mitigated bias in yes ratios $47.43\%$ and $48.66\%$, compared to $44.22\%$ and $47.88\%$ of vanilla DPO. This suggests that V-DPO achieves better hallucination performance while mitigating the over-reliance on language priors with visual guidance.

\paragraph{AMBER.} In Table~\ref{tab:amber}, our approach achieves significant improvements on both AMBER's generative and discriminative tasks. For CHAIR scores, we observe an absolute improvement of $2.2$ from $7.8$ to $5.6$ when applying V-DPO to the human-annotated data RLHF-V. Compared to vanilla DPO, we observe further improvements due to our method on most metrics in both synthetic and human-annotated scenarios. Notably, with only 5K preference pairs collected via synthetic generation, V-DPO outperforms HA-DPO trained on 16K preference pairs
, with an absolute increase of $3.7$ in AMBER score. This indicates the effect of visual guidance in enhancing visual understanding for hallucination mitigation.

\paragraph{HallusionBench.} In Table~\ref{tab:hb}, we use qAcc, fAcc, and aAcc to assess performance on the question-, figure-, and individual-level tasks, respectively\footnote{The GPT-4 evaluation was performed in June 2024.}. We observe a significant improvement in qAcc of V-DPO trained on the synthetic data, with an absolute increase of $9.01\%$ in the accuracy, compared to $4.17\%$ when using RLHF-V for training. One possible explanation for this gap is that the synthetic data mitigates reliance on language priors more efficiently via image-contrast preference learning.

\begin{figure*}
    \centering
    \begin{minipage}[b]{0.32\textwidth}
        \centering
        \small
        \begin{tabular}{lcc}
            \hline
            \textbf{Approach} & \textbf{Hal}$_{\downarrow}$ & \textbf{Score}$_{\uparrow}$  \\
            \hline
            SFT & $0.62$ & $1.97$ \\
            \hline
            \multicolumn{3}{c}{\textbf{Synthetic Augmented Data}} \\
            DPO  & $0.59$ & $2.12$ \\
            V-DPO & $\mathbf{0.53}$\textcolor{forestgreen}{$_{\downarrow 0.09}$} & $\mathbf{2.36}$\textcolor{forestgreen}{$_{\uparrow 0.39}$} \\
            \hline
            \multicolumn{3}{c}{\textbf{RLHF-V}} \\
            DPO  & $0.60$ & $2.08$ \\
            V-DPO & $0.56$\textcolor{forestgreen}{$_{\downarrow 0.06}$} & $2.16$\textcolor{forestgreen}{$_{\uparrow 0.19}$} \\
            \hline
        \end{tabular}
        \captionof{table}{MMHal-Bench results on hallucination rate (Hal) and overall GPT-4 score.}
        \label{tab:mmhal}
    \end{minipage}
    \hfill
    \begin{minipage}[b]{0.67\textwidth}
    \centering
    \includegraphics[width=\textwidth]{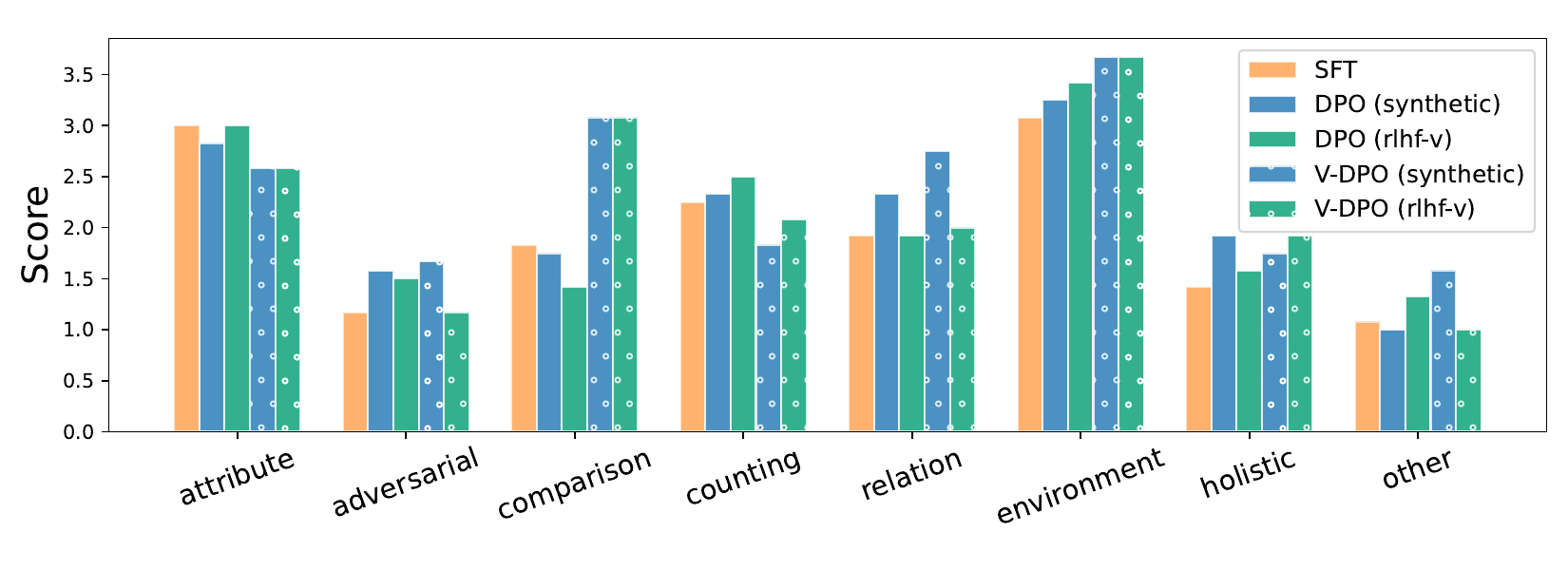}
    \caption{Meso-analysis on MMHal-Bench results comparing performance in different splits of question types.}
    \label{fig:mmhal}
    \end{minipage}
\end{figure*}

\begin{figure*}
    \centering
    \begin{minipage}[b]{0.61\textwidth}
    \centering
    \includegraphics[width=\textwidth]{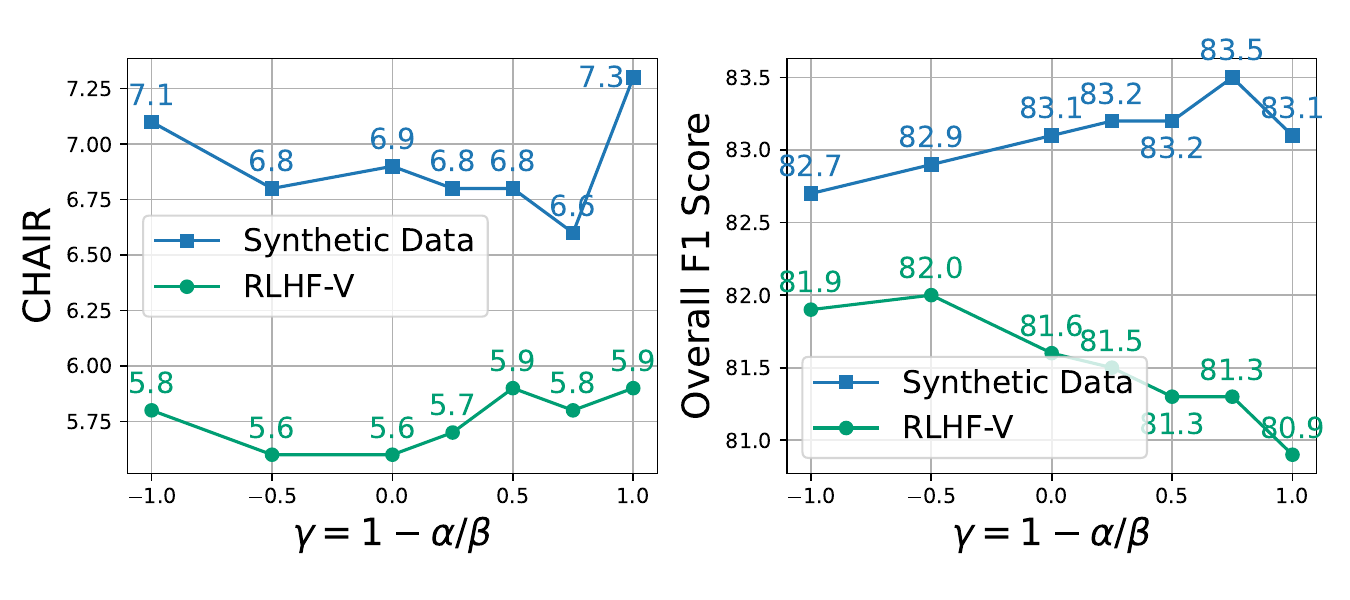}
    \caption{Performance curves (CHAIR$_{\downarrow}$ and F1$_{\uparrow}$) on AMBER with the change of the visual guidance weight $\gamma$.}
    \label{fig:ablation}
    \end{minipage}
    \hfill
    \begin{minipage}[b]{0.37\textwidth}
        \centering
        \small
        \begin{tabular}{lcc}
        \hline
        \textbf{Approach} & \textbf{CHAIR}$_{\downarrow}$ & \textbf{F1}$_{\uparrow}$ \\
        \hline
        \multicolumn{3}{c}{\textbf{Synthetic Augmented Data}} \\
        V-DPO & $6.6$ & $83.5$ \\
        w/ static-lm & $6.3$\textcolor{forestgreen}{$_{\downarrow 0.3}$} & $83.7$\textcolor{forestgreen}{$_{\uparrow 0.2}$} \\
        w/ normalization & $6.2$\textcolor{forestgreen}{$_{\downarrow 0.4}$} & $83.1$\textcolor{red}{$_{\downarrow 0.4}$}  \\
        \hline
        \multicolumn{3}{c}{\textbf{RLHF-V}} \\
        V-DPO & $5.6$ & $81.6$ \\
        w/ static-lm & $5.2$\textcolor{forestgreen}{$_{\downarrow 0.4}$} & $82.4$\textcolor{forestgreen}{$_{\uparrow 0.8}$} \\
        w/ normalization & $5.5$\textcolor{forestgreen}{$_{\downarrow 0.1}$} & $80.4$\textcolor{red}{$_{\downarrow 1.2}$} \\
        \hline
    \end{tabular}
        \captionof{table}{Ablation study on the choice of vision-unconditioned distribution and normalization for V-DPO.}
        \label{tab:ablation}
    \end{minipage}
\end{figure*}

\paragraph{MMHal-Bench.} We conduct GPT-4\footnote{We obtained these results (gpt-4-0613) also in June 2024.} evaluation on MMHal-Bench. Table~\ref{tab:mmhal} presents the hallucination rates and overall scores of the outputs from different models. We observe substantial performance improvements in both synthetic and human-annotated preference data scenarios. Furthermore, we perform meso-analysis on splits of different question types in Figure~\ref{fig:mmhal}. Compared to vanilla DPO, V-DPO is especially effective in answering \textit{comparison} and \textit{environment} questions. Different types of preference data also contribute to the performance gains differently, where our synthetic data shows a superior effect in tackling challenging tasks such as \textit{adversarial} and \textit{relation} questions.

\subsection{Ablation Study}~\label{sec:ablation}
We conduct analyses to investigate the effect of visual guidance in V-DPO. We consider ablations on the $\gamma$-controlled strength of visual guidance, the calculation of vision-unconditioned distribution, and guidance inflation from normalization.

\paragraph{Strength of Visual Guidance.} Figure~\ref{fig:ablation} illustrates the performance changes on AMBER with different values of the visual guidance weight $\gamma$. Specifically, we maintain the same $\beta=0.1$ as in DPO~\citep{DBLP:conf/nips/RafailovSMMEF23} to avoid substantial divergence from the initial model during training and increase $\alpha > 0$ to enhance the strength of visual guidance. When $\gamma=1$, it becomes vanilla DPO without additional enhancement on visual guidance. As $\gamma$ decreases (\textit{i.e.}, $\alpha$ increases), the performance first increases in both training scenarios. However, V-DPO is more sensitive to the guidance control on synthetic preference data, where a small $\gamma$ such as $\gamma = 0.0$\footnote{$\gamma-1=-1$ in Eq.~\ref{eq:vdpo-policy}}
can lead to substantial divergence from the initial model, resulting in performance degradation in hallucination tasks. 
One possible cause of this degradation is the integration of image-contrast data, which may deviate greatly from the initial SFT model generation distributions, increasing the instability of V-DPO given a higher guidance weight. Empirically, we suggest employing data-specific visual guidance control with $\gamma=(0.75,0.00)$ for (synthetic-, human-annotated) scenarios, respectively.

\begin{figure*}
    \centering
    \begin{subfigure}[b]{0.51\columnwidth}
      \includegraphics[width=\textwidth]{figures/vcr-txt.pdf}
      \caption{response-contrast DPO}
      \label{fig:gap-txt-dpo}
  \end{subfigure}
  \hfill
  \begin{subfigure}[b]{0.51\columnwidth}
      \includegraphics[width=\textwidth]{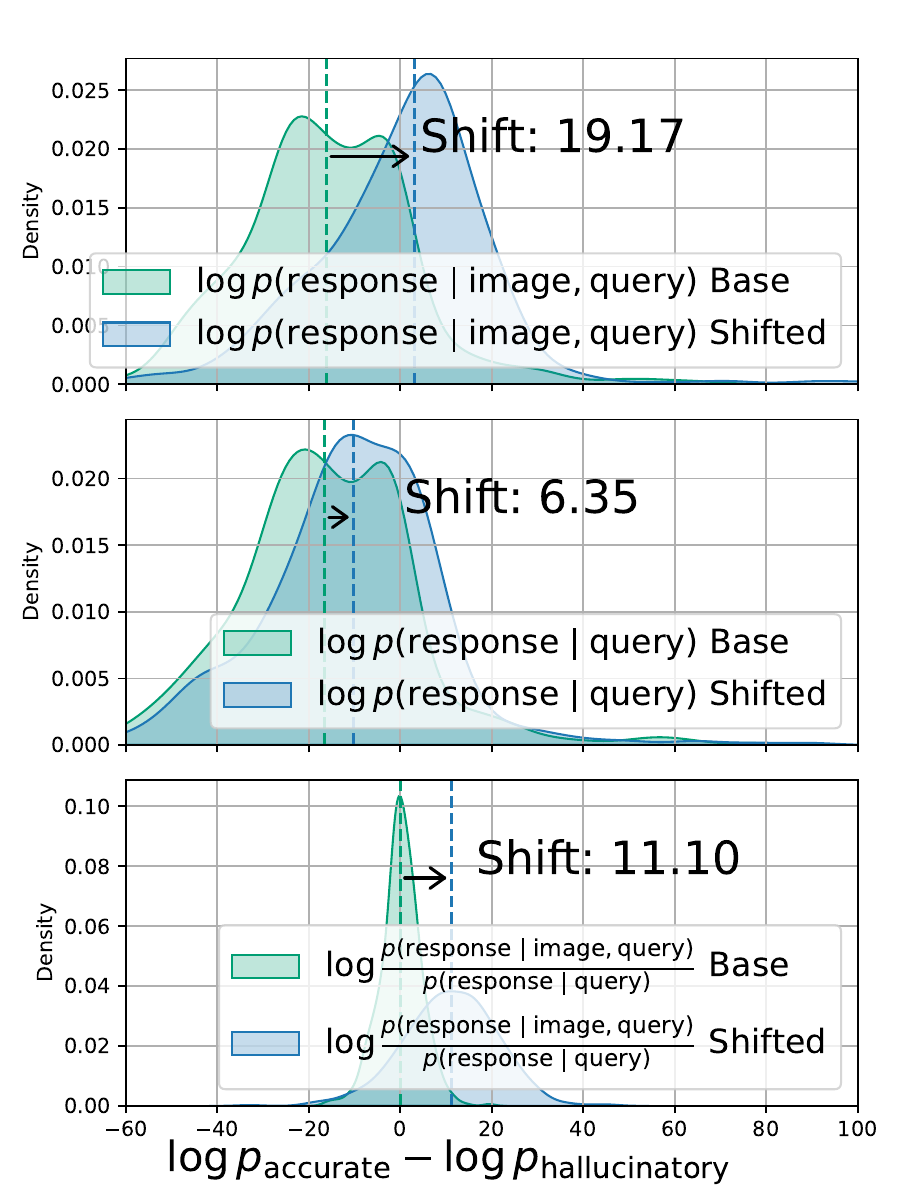}
      \caption{response-contrast V-DPO}
      \label{fig:gap-txt-vdpo}
  \end{subfigure}
  \hfill
  \begin{subfigure}[b]{0.51\columnwidth}
      \includegraphics[width=\textwidth]{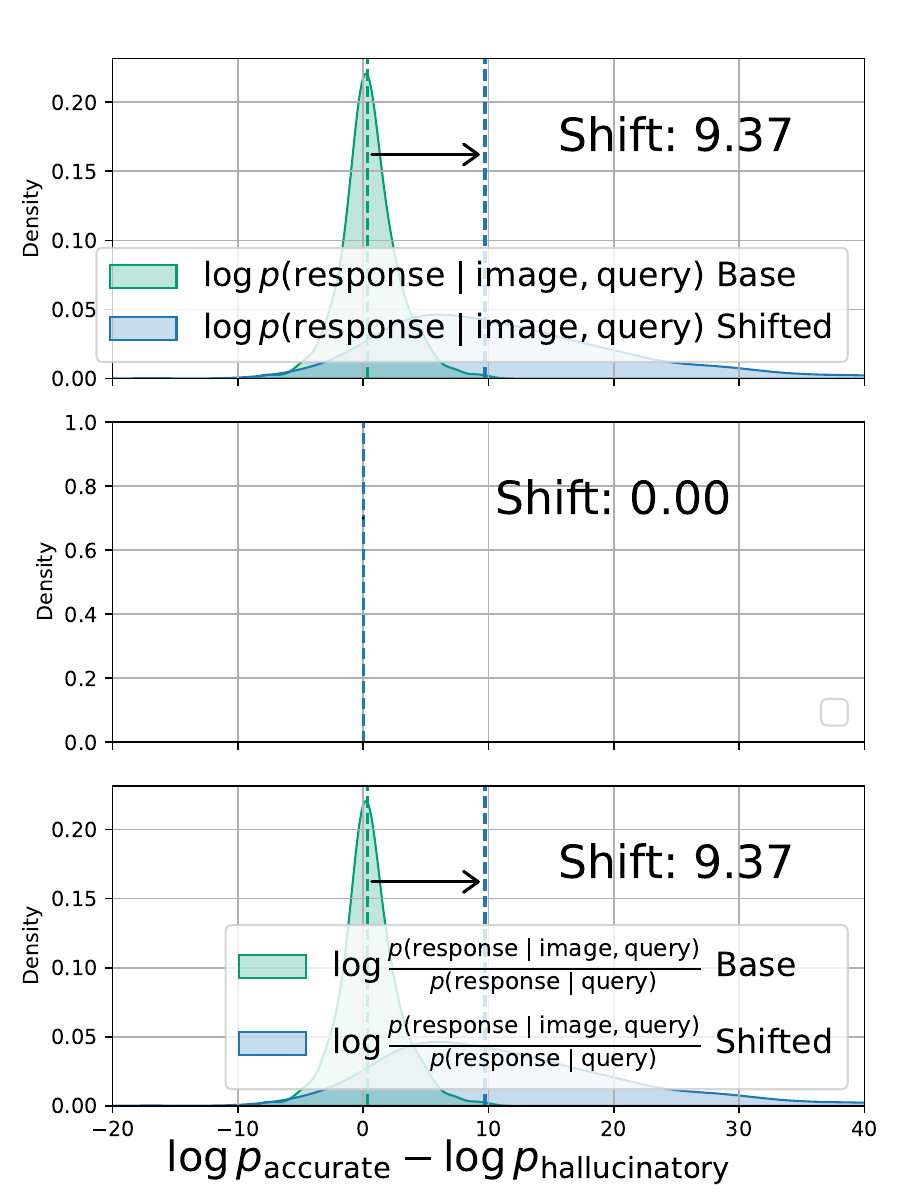}
      \caption{image-contrast DPO}
      \label{fig:gap-txt-dpo}
  \end{subfigure}
  \hfill
  \begin{subfigure}[b]{0.51\columnwidth}
      \includegraphics[width=\textwidth]{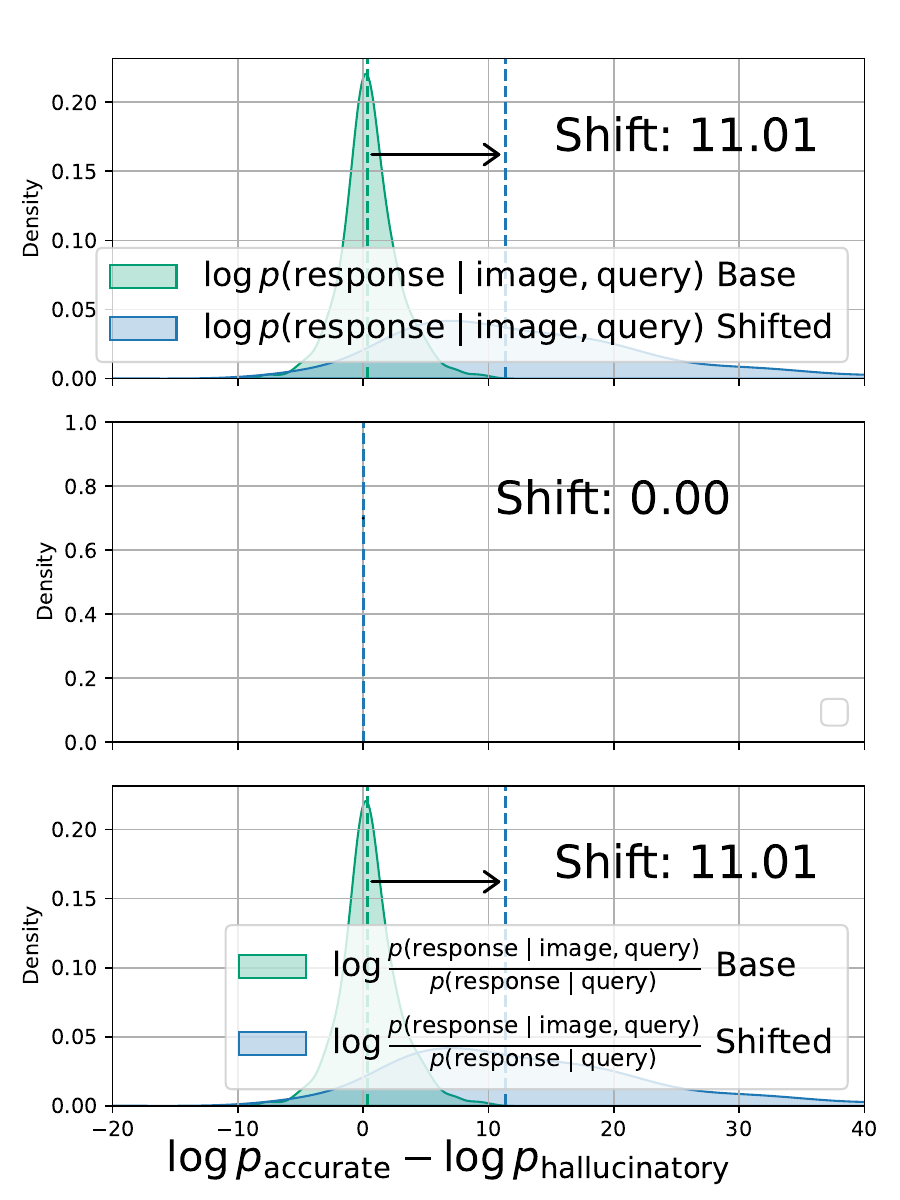}
      \caption{image-contrast V-DPO}
      \label{fig:gap-txt-vdpo}
  \end{subfigure}
  \caption{
  Comparison between V-DPO and vanilla DPO on the shifts of distribution gaps. Rows from top to bottom illustrate the distributions of vision-conditioned generation, textual-only-conditioned generation, and the difference between the two generations, respectively. Note that the shifts equal to $0$ in the textual-only case on image-contrast data, as the samples within a preference pair have the same textual context with each other.}
  \label{fig:gap-all}
\end{figure*}

\begin{figure*}[t]
    \centering
    \includegraphics[width=\linewidth]{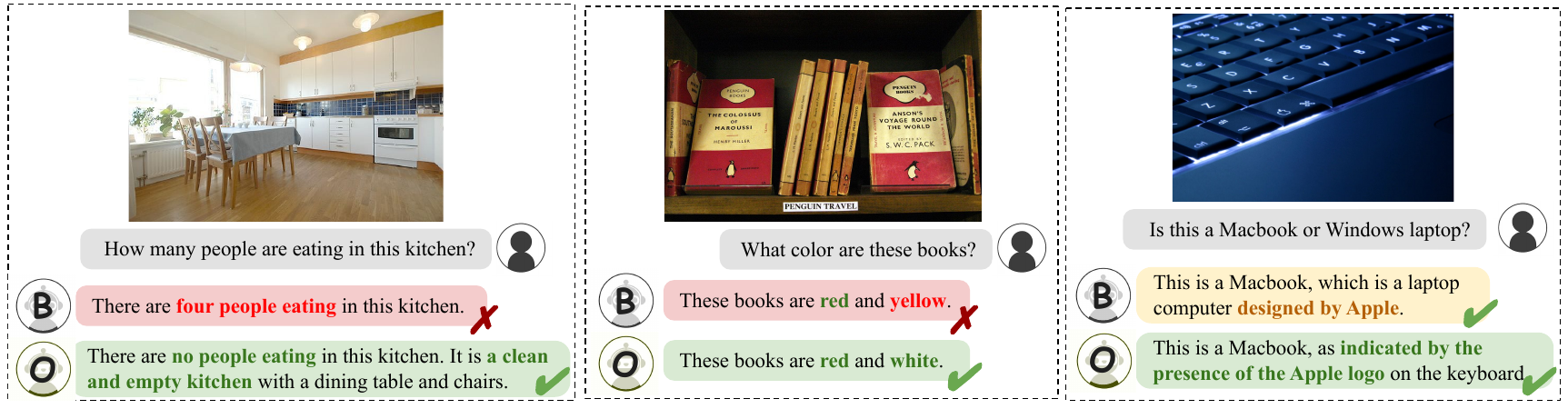}
    \caption{
    Qualitative examples of different predictions of the SFT baseline (\textbf{B}) and our V-DPO approach (\textbf{O}). We \textbf{bolded} keywords indicating the accuracy and informativeness of visual understanding.}
    \label{fig:qualitative}
\end{figure*}

\paragraph{Vision-Unconditioned Distribution Calculation.}
In Eq.~\ref{eq:noimage}, we estimate the vision-unconditioned distribution by replacing the visual representations with zeroes. However, as we only utilize vision-conditioned data for preference learning, the vision-unconditioned distribution can become unreliable due to distribution shifts during training (Figure~\ref{fig:gap}). To interpret the potential influence of the distribution shifts, we use the initial SFT model to calculate the vision-unconditioned distribution instead (\textit{i.e.}, ``w/ static-lm'' in Table~\ref{tab:ablation}). The static textual-only probabilities improve the model performance across both generative and discriminative tasks. This indicates the importance of maintaining reliable vision-unconditioned distribution to integrate appropriate visual guidance during training, shedding light on incorporating textual-only preference data to refine the vision-unconditioned distribution.

\paragraph{Guidance Inflation with Normalization.}
As discussed in Section~\ref{sec:vdpo}, we can normalize the vision-enhanced distribution to inflate the guidance effect. Table~\ref{tab:ablation} shows the model performance after this normalization. Notably, the guidance inflation further mitigates hallucination in generative tasks, achieving lower CHAIR scores (\textit{e.g.}, $6.2$ and $5.5$ compared to $6.6$ and $5.6$) in both data scenarios. However, it may lead to performance drops in discriminative tasks where the result generation distribution is more sensitive to the modified target in preference optimization.

\subsection{Further Analysis}~\label{sec:analysis}
We now investigate the distribution shifts in V-DPO and analyze the qualitative results on MMHal-Bench. Finally, we use the non-hallucination benchmark MMBench to assess the stability of our approach in general tasks in Appendix~\ref{app:mmbench}.

\paragraph{Shifts of Distribution Gaps in V-DPO.} Our ablation study (\S~\ref{sec:ablation}) shows that preference learning can also shift the distribution gaps between accurate and hallucinatory samples in the vision-unconditioned case. In Figure~\ref{fig:gap-all}, we show how V-DPO shifts the distributions across different preference data. Our V-DPO approach is more effective than vanilla DPO in enhancing the ability to determine image-contrast hallucination samples, with a shift of $11.01$, compared with $9.37$ in DPO, as measured by the log-likelihood pairwise preference data differences. For the response-contrast scenario, V-DPO also increases the discriminability with a shift of $19.17$. Furthermore, we observe a smaller shift of $6.35$ in V-DPO in the textual-only distributions compared with that of $7.58$ in DPO, indicating the effectiveness of our approach to mitigate the over-reliance on language priors with visual guidance.

\paragraph{Qualitative Analysis on MMHal-Bench.}
We conduct qualitative analysis to investigate how V-DPO eliminates hallucination in the generated responses. Figure~\ref{fig:qualitative} compares the different generations of V-DPO and the baseline on three examples from MMHal-Bench. The first example, from the adversarial split, shows the significant efficacy of our approach in mitigating the language priors, which may provide a plausible but incorrect answer to the question; \textit{i.e.} ``four people eating''. In the third example, the model learns to justify its answer ``Macbook'' according to the specific visual clue of the ``Apple logo'' in the image. This indicates that our approach enhances visual understanding to elicit related details in the images, improving the informativeness of the generations.

\paragraph{Extended Discussion and Conceptual Comparison.}
As our core focus is the improved DPO variant to enhance visual understanding and alleviate hallucination, we have considered fair comparisons with other DPO-based methods using the same training data or backboned models in our main result tables coding and commonsense reasoning. Specifically, we have included HA-DPO~\citep{DBLP:conf/nips/Dai0LTZW0FH23} which is backboned with the same LVLM (i.e., LLaVA-1.5-7b) on various benchmarks. Other related works on DPO for hallucination mitigation were not included in our fair comparison due to the discrepancy in backboned models (e.g., RLHF-V~\citep{DBLP:journals/corr/abs-2402-11411} on Muffin~\citep{DBLP:conf/iclr/Lou0XSAX0024}). 

For conceptual comparisons, previous works such as POVID~\citep{DBLP:journals/corr/abs-2402-11411} and Silkie~\citep{DBLP:journals/corr/abs-2312-10665} mainly focus on response-contrast data construction to align LVLMs. For example, POVID propose an automatic pipeline to collect dispreferred responses via hallucination injection and image distortion. At the same time, we manipulate the image data to misalign with the original response. Silkie utilizes AI annotation to distill GPT-4V's preference supervision into LVLMs via conventional DPO while we introduce visual-enhanced DPO specifically aimed at LVLM hallucination mitigation. 
\section{Conclusion}~\label{sec:conclusion}
We propose V-DPO, utilizing Classifier-Free Guidance (CFG) to integrate visual guidance in LVLM preference learning. Integrating visual guidance into the training process enhances visual context understanding via preference optimization, improving the accuracy and specificity of model generations. Extensive experiments on various preference data demonstrate the generalizability of V-DPO. We hope our work sheds light on visual guidance for more general tasks in LVLM alignment.

\section*{Limitations}
The main limitations of our work come from two parts. The first one, regarding the V-DPO approach, is the unexplored domains where the language priors are important to guide LVLMs to provide correct answers. For example, preference pairs that prioritize the fluency of the generated text are not considered in our data construction. As this study mainly focuses on the over-reliance on language priors, we leave it to future work to explore more general scenarios where both visual and textual modalities are important to elicit the preferred responses. The second one, related to constructing our synthetic dataset, is the noise and bias introduced by the automatic generation pipeline, which may cause performance degradation during preference optimization. For future work, we may consider a more reliable and scalable way to conduct data filtering and reweighting to refine the quality of synthetic augmented data.

\section*{Ethics Statement}
This work mainly focuses on enhancing visual understanding via preference optimization to mitigate hallucination in LVLMs. One potential ethical concern may come from the data collection process for our synthetic preference pair construction. As the image manipulation process is conducted collaboratively among LVLMs, LLMs, and Stable Diffusion models, systematic bias may be introduced into the generated data. In this case, usage of our synthetic augmented data should be constrained within research-only targets. We leave it to future work to mitigate the bias in model-generated data to further improve the quality of our preference data.

\section*{Acknowledgments}
The authors would like to thank anonymous reviewers for their valuable suggestions. We thank Anirudh Goyal and Michael Shieh for their insightful discussions. We would also like to thank all Web Information Retrieval / Natural Language Processing Group (WING) members for their helpful comments. The computation for this work was partially performed on resources of the National Supercomputing Centre (NSCC), Singapore\footnote{https://www.nscc.sg/}.

\bibliography{custom}
\clearpage
\appendix
\onecolumn
\section{Deriving V-DPO Objective}~\label{app:theory}
Given the maximization objective to optimize in Eq.~\ref{eq:vdpo-obj}, we have:
\begin{equation}
    \label{eq:vdpo-obj-derive}
    \begin{aligned}
        \max_{\pi}\mathbb{E}_{(v,x)\sim\mathcal{I}\times\mathcal{P},y\sim\pi}&\Bigl[r(v,x,y) - \beta\mathbb{D}_{\mathrm{KL}}\left[\pi(y\mid v,x)\parallel\pi_{\mathrm{ref}}(y\mid v,x)\right] + \alpha\mathbb{D}_{\mathrm{KL}}\left[\pi(y\mid v,x)\parallel\pi(y\mid x)\right]\Bigr] \\
        = & \max_{\pi}\mathbb{E}_{(v,x)\sim\mathcal{I}\times\mathcal{P}}\mathbb{E}_{y\sim \pi(y\mid v,x)}\left[r(v,x,y) - \beta\log\frac{\pi(y\mid v,x)}{\pi_{\mathrm{ref}}(y\mid v,x)} + \alpha\log\frac{\pi(y\mid v,x)}{\pi(y\mid x)}\right] \\
        = & \min_{\pi}\mathbb{E}_{(v,x)\sim\mathcal{I}\times\mathcal{P}}\mathbb{E}_{y\sim \pi(y\mid v,x)}\left[\log\frac{\pi(y\mid v,x)}{\pi_{\mathrm{ref}}(y\mid v,x)} - \frac{\alpha}{\beta}\log\frac{\pi(y\mid v,x)}{\pi(y\mid x)} -\frac{1}{\beta}r(v,x,y)\right] \\
        = & \min_{\pi}\mathbb{E}_{(v,x)\sim\mathcal{I}\times\mathcal{P}}\mathbb{E}_{y\sim \pi(y\mid v,x)}\left[\log\frac{\pi(y\mid v,x)^{1-\frac{\alpha}{\beta}}/\pi(y\mid x)^{-\frac{\alpha}{\beta}}}{\frac{1}{Z(v,x)}\pi_{\mathrm{ref}}(y\mid v,x)\exp{\left(\frac{1}{\beta}r(v,x,y)\right)}}-\log{Z(v,x)}\right] \\
        = & \min_{\pi}\mathbb{E}_{(v,x)\sim\mathcal{I}\times\mathcal{P}}\mathbb{E}_{y\sim \pi(y\mid v,x)}\left[\log\frac{\pi(y\mid v,x)\left(\frac{\pi(y\mid v,x)}{\pi(y\mid x)}\right)^{-\frac{\alpha}{\beta}}}{\frac{1}{Z(v,x)}\pi_{\mathrm{ref}}(y\mid v,x)\exp{\left(\frac{1}{\beta}r(v,x,y)\right)}}-\log{Z(v,x)}\right] \\
        = & \min_{\pi}\mathbb{E}_{(v,x)\sim\mathcal{I}\times\mathcal{P}}\mathbb{E}_{y\sim \pi(y\mid v,x)}\left[\log\frac{\pi(y\mid v,x)\left(\frac{\pi(y\mid v,x)}{\pi(y\mid x)}\right)^{\gamma - 1}}{\frac{1}{Z(v,x)}\pi_{\mathrm{ref}}(y\mid v,x)\exp{\left(\frac{1}{\beta}r(v,x,y)\right)}}-\log{Z(v,x)}\right]
    \end{aligned}
\end{equation}
where we set $\gamma=1-\frac{\alpha}{\beta}$ and the partition function is:
$$
    Z(v,x) = \sum_y{\pi_{\mathrm{sft}}(y\mid v,x)\exp{\Bigl(\frac{1}{\beta}r(v,x,y)\Bigl)}}.
$$

Following~\citet{DBLP:conf/nips/RafailovSMMEF23}, we define:
$$
\pi^{*}(y\mid v,x) = \frac{1}{Z(v,x)}\pi_{\mathrm{ref}}(y\mid v,x)\exp\left(\frac{1}{\beta}r(v,x,y)\right)
$$
as a valid normalized probability distribution. Different from vanilla DPO, we have the non-normalized term $\pi(y\mid v,x)\left(\frac{\pi(y\mid v,x)}{\pi(y\mid x)}\right)^{\gamma - 1}$ in our V-DPO objective, which cannot be directly optimized to be $\pi^{*}(y\mid v,x)$. Rearranging Eq.~\ref{eq:vdpo-obj-derive} with normalization, we have:
\begin{equation}
    \label{eq:vdpo-obj-derive2}
    \begin{aligned}
        \min_{\pi}\mathbb{E}_{(v,x)\sim\mathcal{I}\times\mathcal{P}}\mathbb{E}_{y\sim \pi(y\mid v,x)}\left[\log\frac{\frac{1}{W_{\pi}(v,x)}\pi(y\mid v,x)\left(\frac{\pi(y\mid v,x)}{\pi(y\mid x)}\right)^{\gamma - 1}}{\frac{1}{Z(v,x)}\pi_{\mathrm{ref}}(y\mid v,x)\exp{\left(\frac{1}{\beta}r(v,x,y)\right)}}-\log{Z(v,x)} + \log{W_{\pi}(v,x)}\right]
    \end{aligned}
\end{equation}
where the partition function:
$$
W_{\pi}(v,x) = \sum_{y\sim\pi(y\mid v,x)}\pi(y\mid v,x)\left(\frac{\pi(y\mid v,x)}{\pi(y\mid x)}\right)^{\gamma - 1}
$$
depends on the policy $\pi$. Therefore, we cannot directly solve the normalized vision-enhanced probability distribution using $\pi^{*}(y\mid v,x)$. As $\gamma < 1$, $W_{\pi}(v, x)$ decreases when the vision-conditioned distribution diverges from the textual-only one. As the LVLM is aligned with the LLM backbone, we can make the following proposition:
\begin{proposition}
    $\exists M < \infty$, for any $y\sim\pi(y\mid v,x)$, the ratio of $\frac{\pi(y\mid x)}{\pi(y\mid v,x)}$ is bounded by $M$
    \label{prop:1}
\end{proposition}
Proposition~\ref{prop:1} holds, according to the practical observation that the LVLM mainly fits well on the seen image data during training while maintaining a similar distribution with the textual-only generation when given unseen images. Based on proposition~\ref{prop:1}, we take $\min_{\pi}\mathbb{E}\log{W_{\pi}(v,x)}$ as a secondary target and focus on minimizing the first term in Eq.~\ref{eq:vdpo-obj-derive} and~\ref{eq:vdpo-obj-derive2} to elicit an approximation of the optimal solution.

For Eq.~\ref{eq:vdpo-obj-derive}, one straightforward but probably sub-optimal solution is to solve the vision-enhanced distribution with a proportional constraint with $\pi^{*}(y\mid v,x)$:
\begin{equation}
    \label{eq:vdpo-policy2}
    \begin{aligned}
        \pi(y\mid v,x) \left(\frac{\pi(y\mid v,x)}{\pi(y\mid x)}\right)^{\gamma - 1} \varpropto \pi^{*}(y\mid v,x)
    \end{aligned}
\end{equation}

For Eq.~\ref{eq:vdpo-obj-derive2}, we can solve the normalized probability distribution directly using $\pi^{*}(y\mid v,x)$:
\begin{equation}
    \label{eq:vdpo-policy3}
    \begin{aligned}
        \frac{1}{W_{\pi}(v,x)}\pi(y\mid v,x) \left(\frac{\pi(y\mid v,x)}{\pi(y\mid x)}\right)^{\gamma - 1} = \pi^{*}(y\mid v,x)
    \end{aligned}
\end{equation}
Hence, we complete the derivations for Eq.~\ref{eq:vdpo-obj} and \ref{eq:vdpo-policy}.

\section{Implementation Details}\label{app:impl}
We tune the initial SFT model, LLaVA-v1.5-7B, using our V-DPO and the vanilla DPO approaches with the highest learning rate $1$e-$6$ through $4$ epochs on both synthetic and human-annotated data scenarios. We adopt a batch size of $64$ and set $\beta=0.1$, following the DPO paper~\citep{DBLP:conf/nips/RafailovSMMEF23}. We employ different weights of visual guidance on the synthetic ($\gamma=0.75$) and human-annotated ($\gamma=0.0$) data according to their sensitivity to the control strength. All experiments are conducted with a maximum of $4 \times 40$GB GPUs (NVIDIA A100).

\section{More Details in Preference Data Construction}~\label{app:data}
We choose the images from COCO~\citep{DBLP:conf/eccv/LinMBHPRDZ14}, Visual-Genome~\citep{DBLP:journals/ijcv/KrishnaZGJHKCKL17}, Visual Commonsense Reaosning (VCR)~\citep{DBLP:journals/corr/abs-1811-10830} as the seed set for our synthetic data augmentation pipeline, covering various types of visual content including daily-life scenes and drama-event or human-involved scenarios. Our result synthetic augmented data contains $ $ preference pairs, including $ $ image-contrast and $ $ response-contrast samples on visual instruction following, visual question answering, and region description tasks.

\begin{figure}[t]
    \centering
  \begin{subfigure}[b]{0.32\columnwidth}
      \includegraphics[width=\textwidth]{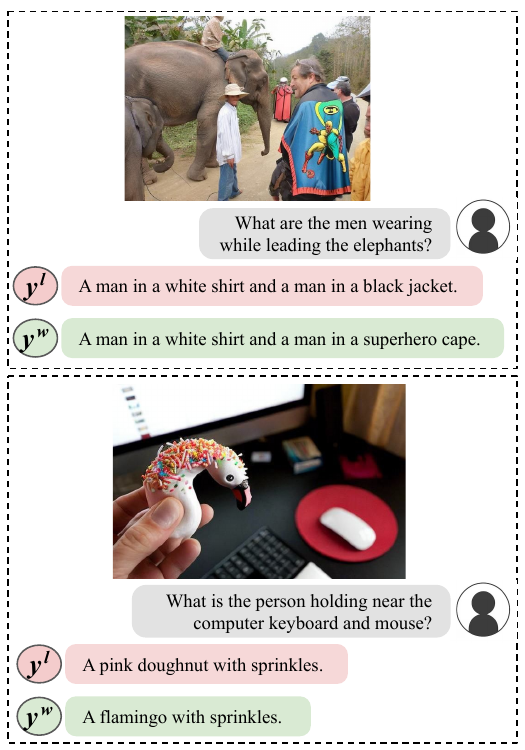}
      \caption{response-contrast vqa}
      \label{fig:txt}
  \end{subfigure}
  \hfill
  \begin{subfigure}[b]{0.32\columnwidth}
      \includegraphics[width=\textwidth]{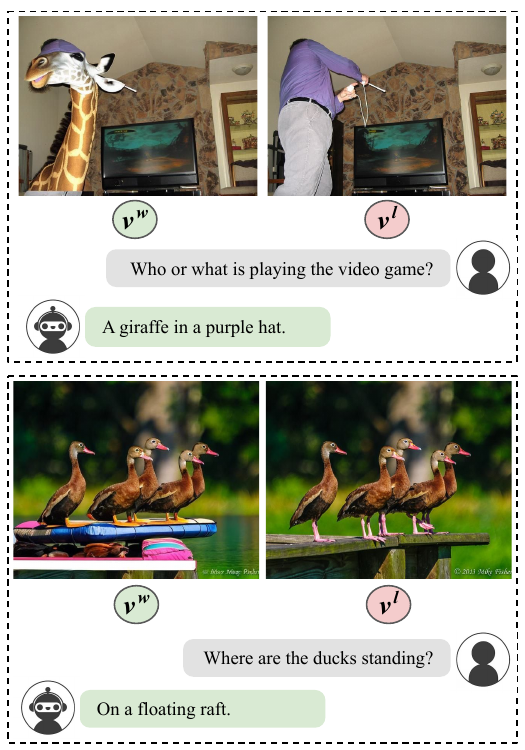}
      \caption{image-contrast vqa}
      \label{fig:img}
  \end{subfigure}
  \hfill
  \begin{subfigure}[b]{0.32\columnwidth}
      \includegraphics[width=\textwidth]{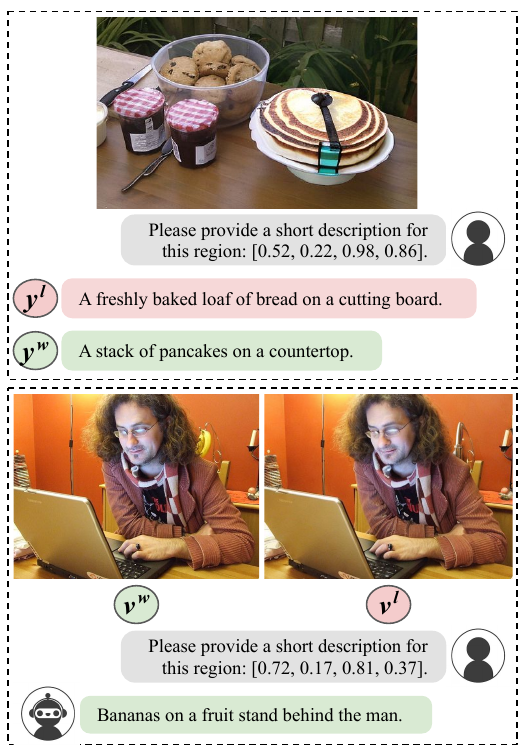}
      \caption{region description}
      \label{fig:reg}
  \end{subfigure}
  \caption{Examples of generated preference data.}
  \label{fig:clip}
\end{figure}
\begin{table}[t]
    \centering
    \small
    \begin{tabular}{l}
        \hline
        \quad\textbf{<< Element Replacement >>} \\
        \textbf{System:} You are a good assistant to help me do academic research. \\
        \textbf{User:} I have an image with the caption: ``\underline{A train is passing by a church.}''. Substitute each of the following objects with \\
        something unexpected to create a sense of discordance: \underline{train, church} in the format: [what] -> [what]. Provide a brief \\
        sentence explaining each substitution. \\
        \textbf{Assistant:} \\
        \hline
        \quad\textbf{<< Captioning for Manipulated Images >>} \\
        \textbf{System:} You are a good assistant to generate new captions. \\
        \textbf{User:} I have an original caption and a substitution operation. Return the new caption after conducting the substitution. \\
        The original caption is: \underline{A train is passing by a church. The substitution involves changing the train to an elephant.} \\
        Return the updated caption. \\
        \textbf{Assistant:} \\
        \hline
        \quad\textbf{<< Question Generation >>} \\
        \textbf{System:} You are a good assistant to generate questions. \\
        \textbf{User:} I have a pair of descriptions. Could you help me generate a question that will lead to different answers based on \\
        the two descriptions? Ensure that the question is suitable for both descriptions. \\
        The first description is: \underline{A woman is cleaning her dining room}. \\
        The second description is: \underline{A robot is cleaning her dining room}. \\
        Return a question and the corresponding answers according to the two descriptions. \\
        \textbf{Assistant:} \\
        \hline
        \quad\textbf{<< Distractor (Answer Candidate) Generation >>} \\
        \textbf{System:} You are a good assistant to generate possible answers. \\
        \textbf{User:} Given a question, please help me to generate some reasonable answers that are common in the real life. \\
        The question is: \underline{Where is the bear sitting?} \\
        A reasonable answer can be: \underline{In a grassy area}. \\
        An unreasonable answer can be: \underline{In a floating jelly beans}. Please help me to generate several reasonable answers, \\
        and seperate each answer with ``|''. \\
        \textbf{Assistant:} \\
        \hline
    \end{tabular}
    \caption{Prompt Templates to utilize LLMs to guide the image manipulation process.}
\label{tab:prompts}
\end{table}

\subsection{Prompts for Image Manipulation}~\label{app:prompts}
We show the designed prompts to elicit element replacement ideas from LLMs such as ChatGPT\footnote{We used gpt-3.5-turbo-1106.}~\citep{DBLP:journals/corr/abs-2303-08774} in Table~\ref{tab:prompts} and examples of generated preference pairs in Figures~\ref{fig:txt} to~\ref{fig:reg}. 

\subsection{Filtering via CLIPScore}~\label{app:clipscore}
Figure~\ref{fig:clip} shows the distributions regarding the difference in CLIPScore between positive and negative samples before filtering. We set a threshold $r=\frac{\mathrm{CLIPScore}^w}{\mathrm{CLIPScore}^l} \ge t = 1.5$ to approve the synthetic samples as a valid preference pair. 

\begin{figure}[h]
  \centering
  \begin{subfigure}[b]{0.45\columnwidth}
      \includegraphics[width=\textwidth]{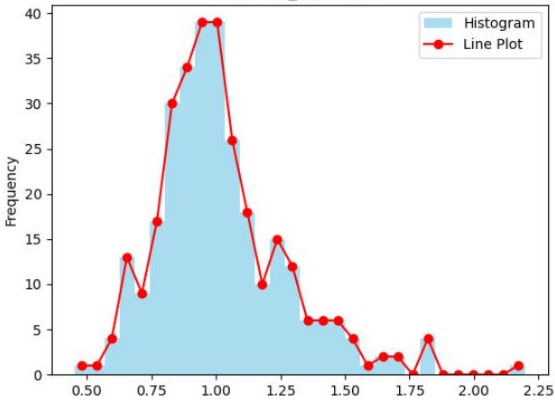}
      \caption{response-contrast}
      \label{fig:clip-txt}
  \end{subfigure}
  \hfill
  \begin{subfigure}[b]{0.45\columnwidth}
      \includegraphics[width=\textwidth]{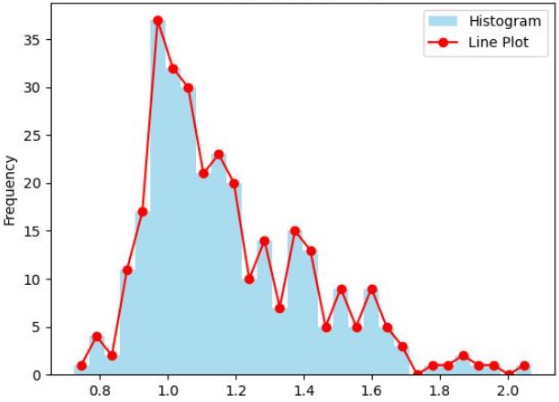}
      \caption{image-contrast}
      \label{fig:clip-img}
  \end{subfigure}
  \caption{Distributions of CLIPScore ratios of unfiltered generated preference pairs.}
  \label{fig:clip}
\end{figure}

\section{Extended Experiments}\label{app:exp}
In this section, we demonstrate the generalizability of V-DPO with extended experiments on \textsc{LLaVA-1.6-vicuna-7b}. We observe that both our method (V-DPO) and conventional DPO significantly reduce hallucination on AMBER (e.g., from $86.7$ to $88.9$ when trained with synthetic data using V-DPO), while maintaining comparable performance on the general evaluation benchmark MMBench in Table~\ref{tab:mmbench}.
\begin{table*}
  \centering
  \small
  \begin{tabular}{l cccc cccc c}
    \hline
    \multirow{2}{*}{\textbf{Approach}} & \multicolumn{4}{c}{\textbf{Generative}} & \multicolumn{4}{c}{\textbf{Discriminative}} & \multirow{3}{*}{\shortstack{\textbf{AMBER}\\\textbf{Score}$_{\uparrow}$}} \\
    \cmidrule(lr){2-5} \cmidrule(lr){6-9}
    & \textbf{CHAIR$_{\downarrow}$} & \textbf{Cover$_{\uparrow}$} & \textbf{Hal$_{\downarrow}$} & \textbf{Cog$_{\downarrow}$} & \textbf{F1\textsubscript{E}$_{\uparrow}$} & \textbf{F1\textsubscript{A}$_{\uparrow}$} & \textbf{F1\textsubscript{R}$_{\uparrow}$} & \textbf{F1$_{\uparrow}$} &  \\
    \hline
    SFT & $8.8$ & $58.4$ & $47.8$ & $4.2$ & $90.4$ & $74.3$ & $69.4$ & $82.3$ & $86.7$ \\
    \hline 
    \multicolumn{10}{c}{\textbf{Synthetic Augmented Data}} \\
    DPO & $8.8$ & $57.3$ & $47.5$ & $3.7$ & $95.6$ & $77.4$ & $64.3$ & $84.9$ & $88.0$ \\
    V-DPO (Ours) & $8.5$\textcolor{forestgreen}{$_{\downarrow 0.3}$} & $56.2$\textcolor{red}{$_{\downarrow 1.1}$} & $46.6$\textcolor{forestgreen}{$_{\downarrow 1.1}$} & $3.5$\textcolor{forestgreen}{$_{\downarrow 0.2}$} & $97.3$ & $\mathbf{77.7}$ & $64.6$ & $\mathbf{86.3}$\textcolor{forestgreen}{$_{\uparrow 1.4}$} & $\mathbf{88.9}$\textcolor{forestgreen}{$_{\uparrow 0.9}$} \\
    \hline
    \multicolumn{10}{c}{\textbf{RLHF-V}} \\
    DPO & $8.7$ & $57.5$ & $\mathbf{46.3}$ & $\mathbf{3.9}$ & $94.2$ & $77.2$ & $68.6$ & $85.0$ & $88.1$ \\
    V-DPO (Ours) & $\mathbf{8.4}$\textcolor{forestgreen}{$_{\downarrow 0.3}$} & $57.4$\textcolor{red}{$_{\downarrow 0.1}$} & $\mathbf{42.3}$\textcolor{forestgreen}{$_{\downarrow 3.0}$} & $3.5$\textcolor{forestgreen}{$_{\downarrow 0.4}$} & $95.8$ & $77.2$ & $68.6$ & $85.9$\textcolor{forestgreen}{$_{\uparrow 0.9}$} & $88.7$\textcolor{forestgreen}{$_{\uparrow 0.7}$} \\
    \hline
  \end{tabular}
  \caption{Result comparison on AMBER. We compare methods backboned with LLaVA-v1.6-Vicuna-7B.}
  \label{tab:amber-1.6}
\end{table*}

\section{General Evaluation on MMBench}\label{app:mmbench}
One drawback of alignment methods is the enlarged divergence from the initial SFT model through training, potentially resulting in model performance degradation on general multimodal tasks. Table~\ref{tab:mmbench} assesses V-DPO on the general evaluation benchmark MMBench. While V-DPO still causes a slight drop in overall accuracy, we observe a relatively improved performance compared to the vanilla DPO on both synthetic and human-annotated data scenarios. We leave it to future work to further enhance the stability and generalizability of V-DPO across more general tasks in LVLMs.

\begin{table*}[h]
    \centering
    \small
    \begin{tabular}{lcccccccc}
            \hline
            \multirow{2}{*}{\textbf{Approach}} & \multirow{2}{*}{\textbf{Model}} & \multicolumn{6}{c}{\textbf{Level-2 Capability Accuracy}} & \multirow{3}{*}{\shortstack{\textbf{Overall}\\\textbf{Accuracy}$_{\uparrow}$}} \\
            \cmidrule(lr){3-8}
            && \textbf{AR}$_{\uparrow}$ & \textbf{CP}$_{\uparrow}$ & \textbf{FP-C}$_{\uparrow}$ & \textbf{FP-S}$_{\uparrow}$ & \textbf{LR}$_{\uparrow}$ & \textbf{RR}$_{\uparrow}$ &  \\
            \hline
            SFT & \textsc{llava-1.5-7b} & $73.37$ & $77.70$ & $57.34$ & $68.94$ & $32.20$ & $53.04$ & $65.21$ \\
            SFT & \textsc{llava-1.6-7b} & $72.36$ & $79.05$ & $56.64$ & $67.92$ & $38.13$ & $\mathbf{60.87}$ & $\mathbf{66.41}$ \\
            \hline
            \multicolumn{9}{c}{\textbf{Synthetic Augmented Data}} \\
            \multirow{2}{*}{DPO} & \textsc{llava-1.5-7b} & $\mathbf{74.37}$ & $76.35$ & $56.64$ & $\mathbf{68.94}$ & $32.20$ & $53.91$ & $65.03$ \\
             & \textsc{llava-1.6-7b} & $70.85$ & $79.05$ & $56.64$ & $67.24$ & $38.14$ & $60.00$ & $65.89$ \\
            \multirow{2}{*}{V-DPO} &  \textsc{llava-1.5-7b}& $\mathbf{74.37}$ & $76.01$ & $\mathbf{58.04}$ & $\mathbf{68.94}$ & $31.36$ & $54.78$ & $65.12$ \\
             & \textsc{llava-1.6-7b} & $70.85$ & $79.05$ & $57.23$ & $67.58$ & $\mathbf{38.98}$ & $60.00$ & $66.15$ \\
            \hline
            \multicolumn{9}{c}{\textbf{RLHF-V}} \\
            \multirow{2}{*}{DPO} & \textsc{llava-1.5-7b} & $\mathbf{74.37}$ & $76.01$ & $57.34$ & $68.60$ & $31.36$ & $53.04$ & $64.78$ \\
              & \textsc{llava-1.6-7b} & $70.85$ & $78.72$ & $56.64$ & $66.55$ & $36.44$ & $\mathbf{60.87}$ & $65.55$ \\
            \multirow{2}{*}{V-DPO} & \textsc{llava-1.5-7b} & $73.87$ & $76.69$ & $57.34$ & $68.60$ & $32.20$ & $53.04$ & $64.95$ \\
              & \textsc{llava-1.6-7b} & $70.85$ & $\mathbf{79.73}$ & $56.64$ & $67.24$ & $38.14$ & $60.00$ & $66.07$ \\
            \hline
    \end{tabular}
    \caption{MMBench results}
    \label{tab:mmbench}
\end{table*}

\end{document}